\documentclass{article} % For LaTeX2e
\usepackage{iclr2027_conference,times}

\usepackage{amsmath,amsfonts,bm}

\def\eqref#1{equation~\ref{#1}}
\def\1{\bm{1}}

\DeclareMathAlphabet{\mathsfit}{\encodingdefault}{\sfdefault}{m}{sl}
\SetMathAlphabet{\mathsfit}{bold}{\encodingdefault}{\sfdefault}{bx}{n}

\usepackage{hyperref}
\usepackage{url}
\usepackage[ruled,vlined]{algorithm2e}
\usepackage{lmodern}
\usepackage{float}
\usepackage[table]{xcolor}
\usepackage{multirow}
\usepackage{graphicx} 
\usepackage{enumitem}
\usepackage{booktabs}
\usepackage{tabularx}
\usepackage{array}
\usepackage{wrapfig}
\usepackage{subcaption}

\let\titleold\title
\renewcommand{\title}[1]{\titleold{#1}\newcommand{\thetitle}{#1}}
\def\maketitlesupplementary
{
    \newpage
    \begin{center}
        \Large
        \textbf{\thetitle}\\
        \vspace{0.5em}Supplementary Material \\
        \vspace{1.0em}
    \end{center}
}

\title{MixVLA: Adaptive Mixing of Non-Invariant Information for Generalizable Vision-Language-Action Models}

\author{Pingrui Zhang$^{1,2}$, Yu Zhang$^{3}$, Pengyuan Wu$^{2}$, Bin Wang$^{2}$, Haoming Song$^{2}$, \\
\textbf{Xianqiang Gao}$^{2}$, \textbf{ZhaxiZhuoma}$^{2}$, \textbf{Zhigang Wang}$^{2}$, \textbf{Dong Wang}$^{2}$, \textbf{Bin Zhao}$^{2,4}$,\\ \textbf{Xuelong Li}$^{5 \dagger}$\\
\textsuperscript{1}Fudan University \quad
\textsuperscript{2}Shanghai AI Laboratory \\
\textsuperscript{3}Harbin Institute of Technology, Shenzhen, China \\
\textsuperscript{4}Northwestern Polytechnical University \quad
\textsuperscript{5}TeleAI, China Telecom Corp Ltd \\
    \texttt{zhangpingrui@pjlab.org.cn \quad xuelong\_li@ieee.org} \\
}

\newcommand{\ours}{MixVLA\xspace}

\definecolor{tabfirst}{RGB}{189, 201, 225}
\definecolor{tabsecond}{RGB}{224, 230, 240}
\definecolor{tabthird}{rgb}{1, 1, 0.7}

\newcommand{\cfirst}{\cellcolor{tabfirst}\bfseries}
\newcommand{\csecond}{\cellcolor{tabsecond}}

\iclrfinalcopy

\begin{document}

\maketitle

\begin{abstract}
Vision-Language-Action (VLA) models have achieved remarkable advances in robotic manipulation, yet their zero-shot generalization under out-of-distribution (OOD) conditions remains limited. 
These models often entangle task-relevant invariant structure with environment-specific non-invariant factors, causing policies to rely on spurious appearance cues during action prediction.
In this work, we propose \textbf{MixVLA}, a model-agnostic training framework that improves the generalization of VLA models without requiring additional OOD data or architectural modifications. 
The key component of MixVLA is \textbf{Adaptive Mixing of Non-Invariant Information (AMI)}. 
AMI stochastically mixes non-invariant representations to regularize distribution-specific variability while preserving complementary predictive cues. 
The mixed non-invariant features are then fused with invariant representations for final action prediction, resulting in improved robustness without sacrificing policy expressiveness. 
Extensive experiments across challenging manipulation settings, including LIBERO, LIBERO-Plus, the RoboTwin perturbation suite, and real-world tasks, demonstrate that MixVLA improves overall zero-shot robustness while retaining strong in-domain performance.
\end{abstract}

\section{Introduction}
\label{sec:intro}

Vision-Language Models (VLMs)~\citep{achiam2023gpt,bai2023qwen} have advanced rapidly in recent years and are increasingly applied to robotic manipulation.  
This trend has led to the emergence of Vision-Language-Action (VLA) models~\citep{kim2024openvla,brohan2022rt} that map multimodal observations and language instructions into low-level control actions. 
Benefiting from rich cross-modal representations learned by VLMs, these models achieve competitive task success rates across both simulated~\citep{liu2023libero,chen2025robotwin} and real-world~\citep{walke2023bridgedata} manipulation benchmarks. 

\begin{figure*}[t]
\centering
% \vspace{-20pt}
\includegraphics[width=0.99\linewidth]{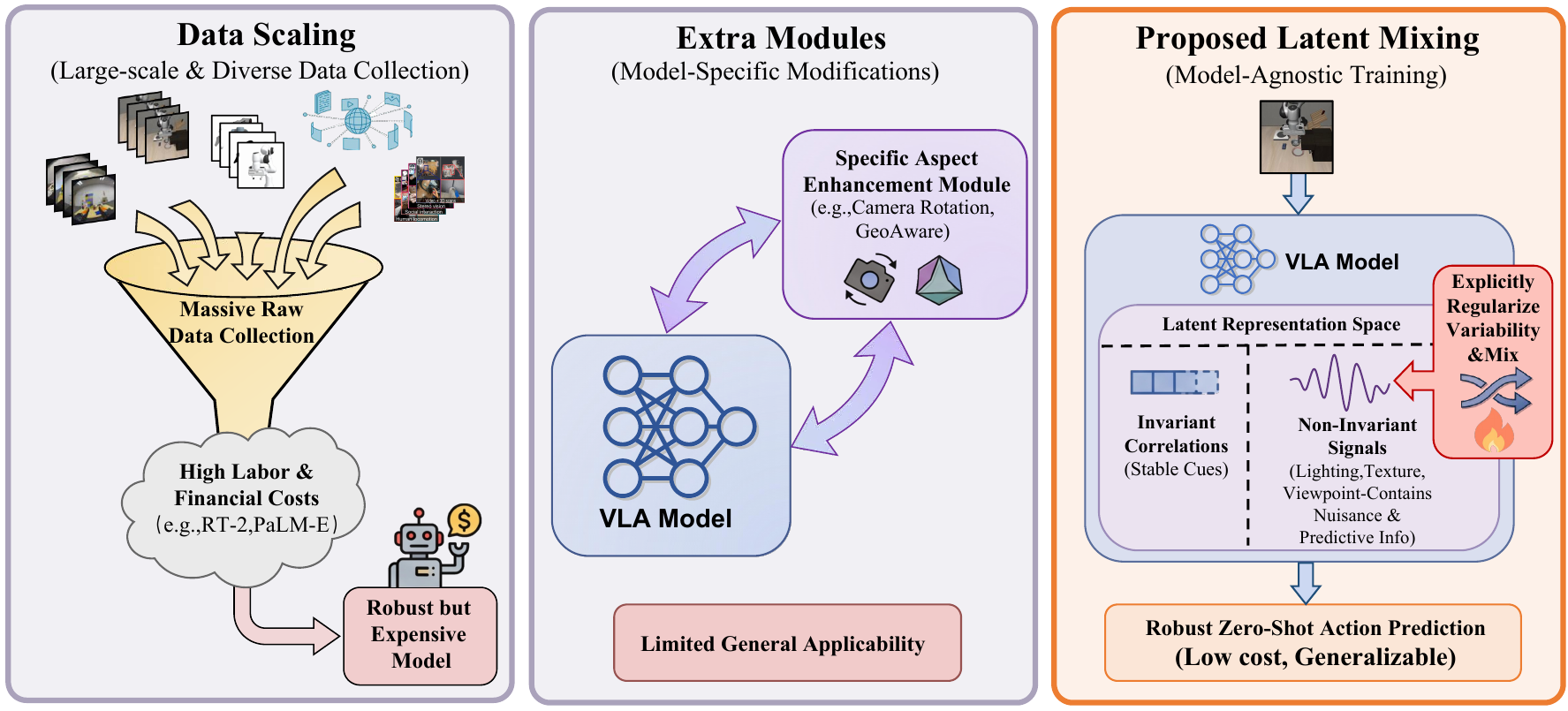}
\vspace{-5pt}
\caption{
\textbf{Left:} Data scaling improves robustness by collecting massive and diverse demonstrations, but incurs substantial labor and financial costs.
\textbf{Middle:} Some works improve generalization with task-specific modules (e.g., camera rotation or geometric awareness), which limits general applicability across VLA architectures.
\textbf{Right:} MixVLA is a model-agnostic training framework that explicitly regularizes and mixes non-invariant signals in the latent representation space, enabling robust zero-shot action prediction without additional data or architectural modifications. 
}
\vspace{-12pt}
\label{fig:motivation}
\end{figure*}

Despite strong performance on in-distribution benchmarks, VLA models still struggle to achieve robust zero-shot generalization under out-of-distribution (OOD) shifts~\citep{li2025cronusvla,zhou2025libero,wang2026libero}. 
Recent studies~\citep{fei2025libero} show that even mild variations in lighting, appearance, or camera viewpoint can cause substantial performance degradation.  
Some prior works~\citep{zitkovich2023rt,generalist2025gen0} attempt to improve robustness through large-scale and diverse data collection, but they incur substantial labor and financial costs. 
Other approaches~\citep{Abouzeid2025GeoAwareVLAIG,Wu2024AffordDPGD} enhance specific aspects of generalization by introducing model-specific modifications, which often limit their general applicability. 
In this work, we instead focus on improving zero-shot generalization through a model-agnostic training strategy that does not rely on additional out-of-distribution data. 
As illustrated in figure~\ref{fig:motivation}, our key idea is to explicitly regularize the variability of non-invariant signals in the latent representation space.

To better understand this intuition, we analyze the problem from a representation learning perspective.
Policies trained with empirical risk minimization (ERM)~\citep{erm} tend to capture both task-relevant invariant correlations and environment-specific non-invariant signals. 
While invariant features provide stable cues for action prediction, non-invariant factors such as lighting, texture, and viewpoint often contain both nuisance variations and complementary predictive information. 
Under ERM training, directly relying on raw non-invariant features can introduce spurious dependencies under distribution shifts. 
Conversely, enforcing strict invariance may be overly conservative, as it can discard informative signals that remain useful for action prediction. 
This observation motivates explicitly regularizing the variability of non-invariant signals rather than relying on invariant features alone.

Building on the above insights, we propose \textbf{MixVLA}, a model-agnostic training framework that improves zero-shot generalization of Vision-Language-Action models, without modifying model architectures or requiring additional data.
The core of MixVLA is \textbf{A}daptive \textbf{M}ixing of Non-\textbf{I}nvariant Information (\textbf{AMI}). 
Specifically, we first disentangle invariant and non-invariant representations using an information-theoretic objective. 
We then perform stochastic mixing within the non-invariant feature space to regularize environment-specific variability while preserving complementary predictive cues. 
The resulting mixed non-invariant representation is subsequently fused with invariant features for final action prediction, enabling a more favorable balance between robustness and policy expressiveness.
We evaluate MixVLA across a range of challenging manipulation settings, including the LIBERO and LIBERO-Plus benchmarks, the RoboTwin perturbation suite, and real-world robotic manipulation tasks.
The results show that MixVLA substantially improves the robustness of VLA policies under diverse observation shifts without requiring additional data collection.

\noindent Our key contributions are three-fold:
\begin{itemize}[leftmargin=*, labelsep=0.5em, topsep=3pt, itemsep=2pt, parsep=0pt]
    \item\textbf{A model-agnostic framework for generalized VLA models.}
    We present \textbf{MixVLA}, a model-agnostic training framework that improves the zero-shot generalization of Vision-Language-Action models under out-of-distribution (OOD) data without additional data collection or architectural redesign. 

    \item\textbf{Adaptive mixing of non-invariant information (AMI).}
    We propose \textbf{AMI}, which adaptively mixes non-invariant representations and fuses them with invariant features to improve robustness under distribution shifts.

    \item\textbf{Consistent zero-shot gains on challenging manipulation benchmarks.}
    Experiments on LIBERO, LIBERO-Plus, RoboTwin, and real-world manipulation tasks demonstrate that MixVLA consistently improves zero-shot task success rates under diverse perturbations.
\end{itemize}

\section{Related Work}

\subsection{Vision-Language-Action Model}

Recent Vision-Language-Action (VLA) models build upon large-scale Vision-Language Models (VLMs)~\citep{bai2023qwen,team2025qwen3,achiam2023gpt,comanici2025gemini,zhang2026evaluating,zhang2026instruction} and are fine-tuned on diverse robot trajectory datasets to learn language-conditioned visuomotor policies~\citep{brohan2022rt, zitkovich2023rt, walke2023bridgedata, Ebert2021BridgeDB, o2024open}. 
RT-1 and RT-2~\citep{brohan2022rt, zitkovich2023rt} demonstrate that scaling cross-modal pretraining together with large robot datasets can substantially improve manipulation performance and semantic grounding. 
Following this paradigm, OpenVLA~\citep{kim2024openvla} discretizes robot actions into rarely used vocabulary tokens and performs autoregressive action generation on top of the Prismatic-7B backbone~\citep{Karamcheti2024PrismaticVI}. 
The $\pi_0$ family~\citep{Black20240AV, Intelligence202505AV} further introduces a dual-system design that augments a VLM backbone with a dedicated action expert to jointly model high-level reasoning and low-level control.
Beyond architectural scaling, recent efforts explore richer spatial representations~\citep{Zhen20243DVLAA3, Bhat20253DCL, zhang20254dvlaspatiotemporalvisionlanguageactionpretraining,zhang2026cross}, intermediate supervision such as subtask decomposition and trajectory prediction~\citep{Zheng2024TraceVLAVT, Zhao2025CoTVLAVC, Zhang2025DreamVLAAV, zhong2025flowvlavisualchainthoughtbased, Cen2025WorldVLATA, Lee2025MolmoActAR,zhang2025moma}, and post-training optimization via planning or reinforcement learning~\citep{Guo2025VLAReasonerEV, zhang2025grapegeneralizingrobotpolicy, Lu2025VLARLTM, Fei2025SRPOSP, Huang2025ThinkActVR} to enhance VLA capability. 
Despite these advances in scaling and architectural design, improving robustness to observation variations in unseen environments remains an open challenge for VLA policies. This motivates our focus on enhancing the intrinsic generalization ability of VLA models through a model-agnostic training strategy.

\subsection{Generalization in Robotic Manipulation}

Recent work has explored multiple directions to improve generalization in robotic manipulation. 
One common line of work~\citep{brohan2022rt,driess2023palm,zitkovich2023rt,zhaxizhuoma2025fastumi,generalist2025gen0} seeks to enhance robustness through large-scale and diverse data collection. 
By expanding the coverage of training distributions, these approaches aim to improve policy generalization across varied environments. 
However, such strategies often incur substantial human and financial costs, and the practical boundary of data-driven generalization for embodied models remains insufficiently understood~\citep{generalist2025gen0}. 
Complementary to data scaling, another line of research improves robustness by introducing additional architectural modules or auxiliary components. 
For example, GeoAware-VLA~\citep{Abouzeid2025GeoAwareVLAIG} incorporates VGGT~\citep{wang2025vggt} to improve robustness to camera viewpoint changes. 
AffordDP~\citep{Wu2024AffordDPGD} enhances cross-scene manipulation through affordance retrieval and transfer, while 3DS-VLA~\citep{pmlr-v305-li25g} improves robustness to appearance and viewpoint variations by introducing a 3D point cloud perception module. 
While these designs can improve robustness under specific perturbations, they often introduce additional system complexity and may not generalize well to diverse out-of-distribution scenarios, where realistic shifts are typically multifaceted rather than isolated.
A separate line of work~\citep{yang2024equibotsim3equivariantdiffusionpolicy, yang2024equivactsim3equivariantvisuomotorpolicies, Seo_2024, ryu2023equivariantdescriptorfieldsse3equivariant} seeks to improve policy generalization by learning robust visual representations or incorporating structured equivariant priors. 
These approaches typically integrate mechanisms such as associative memory or geometric equivariance to maintain policy consistency under object-level transformations, including scaling, rotation, and translation. 
However, such methods primarily focus on object-centric variations and do not explicitly address variability arising from environment-level nuisance factors, such as lighting changes, background appearance, or camera viewpoint shifts.

In contrast to these directions, our work improves VLA robustness through a model-agnostic training strategy that explicitly regularizes non-invariant environmental variability without requiring additional data or specialized architectural modules.

\section{Method}
\label{sec:method}

\begin{figure*}[t]
\centering
\includegraphics[width=0.99\linewidth]{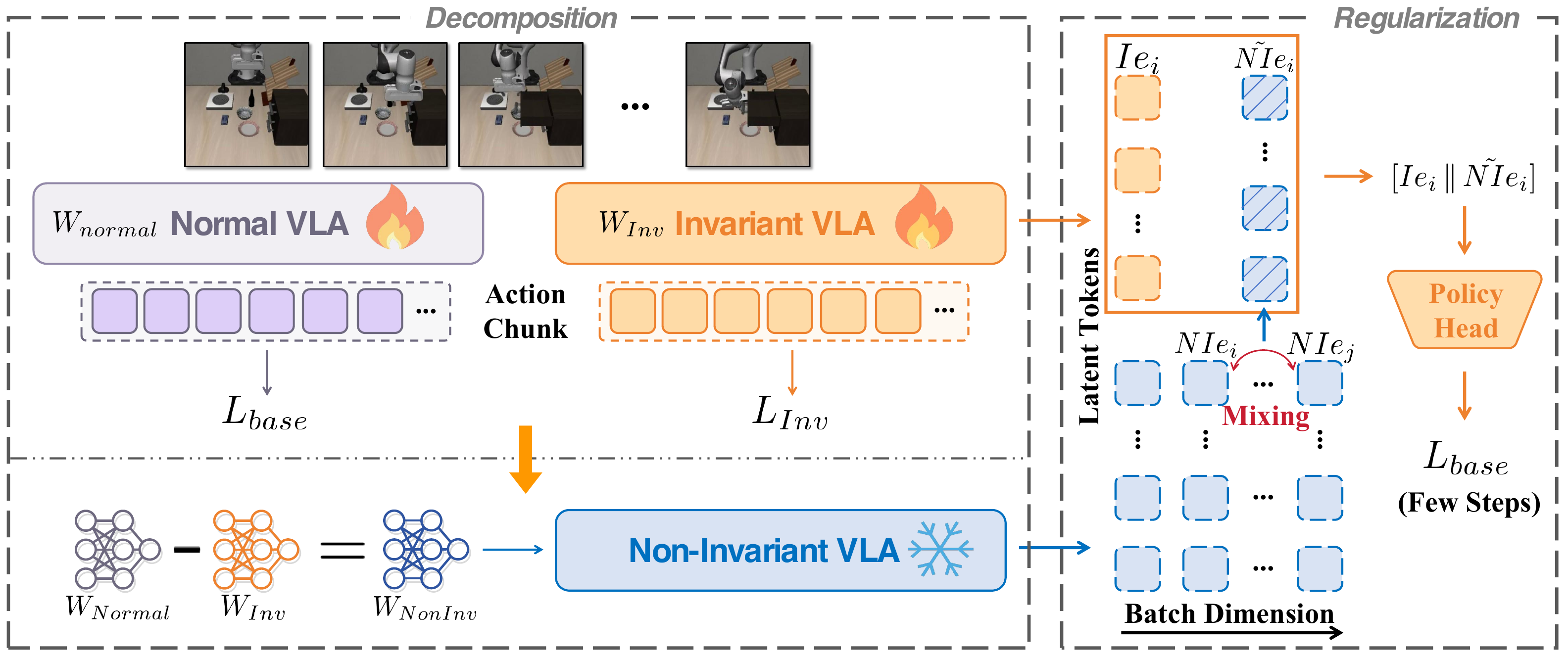}
\vspace{-5pt}
\caption{\textbf{Overview of MixVLA.}
\textbf{Left (Decomposition):} 
A standard VLA policy is trained with the base objective to obtain $W_{\text{Normal}}$, while an Invariant VLA is trained with an information bottleneck objective to obtain $W_{\text{Inv}}$. 
The Non-Invariant VLA is then derived via weight arithmetic.
\textbf{Right (Regularization):} 
Non-invariant embeddings are randomly mixed within the mini-batch using \textbf{AMI}. 
The mixed representation $\tilde{NIe}$ is concatenated with the invariant embedding $Ie$ and passed to the policy head for training.
}
\vspace{-5pt}
\label{fig:overview}
\end{figure*}

\subsection{Preliminaries}
\label{sec:prelim}

\paragraph{VLA policy.} 
We consider a VLA policy that maps an observation–instruction pair to robot actions. 
Let $X$ denote the visual observation and $L$ the language instruction. 
A standard VLA backbone produces a latent representation
$
Z = f_{\theta}(X, L),
$
which is subsequently consumed by the policy head $\pi(\cdot)$ to predict the action $A$. 
MixVLA operates at this latent level without modifying the backbone or the action head.

\paragraph{Invariant and non-invariant factors.}
We hypothesize that $Z$ entangles task-relevant invariant structure shared across environments with environment-dependent non-invariant factors.
MixVLA learns an \emph{Invariant Net} and derives a complementary \emph{Non-Invariant Net} to extract the corresponding embeddings $Ie$ and $NIe$, respectively.
Adaptive mixing regularizes $NIe$ into $\tilde{NIe}$, yielding the fused representation
$
h = [Ie \,\|\, \tilde{NIe}],
$
which is fed to the original policy head for action prediction.
We denote the parameters of the Invariant, Normal, and Non-Invariant Nets by $W_{\text{Inv}}$, $W_{\text{Normal}}$, and $W_{\text{NonInv}}$, respectively (see figure~\ref{fig:overview}).

\subsection{Invariant and Non-Invariant Decomposition}
\label{sec:decomp}

\paragraph{Invariant Net via Information Bottleneck.}
The Invariant Net aims to extract task-critical features that remain stable across environments while suppressing redundant correlations.
We adopt the Information Bottleneck (IB) principle to learn $Ie$ by minimizing $I(X; Ie)$ while maximizing $I(Ie; A)$, balancing input compression and action predictability.
Using the identity
\begin{equation}
I(Ie;A) = H(A) - H(A|Ie),
\end{equation}
to maximize $I(Ie;A)$ is equivalent to minimizing the conditional entropy $H(A|Ie)$, which corresponds to improving action predictability.
Combining sufficiency and minimality yields the classical IB objective:
\begin{equation}
\mathcal{L}_{\text{IB}}
=
\beta I(X;Ie) - I(Ie;A).
\end{equation}
In practice, we optimize the following surrogate:
\begin{equation}
\mathcal{L}_{\text{Inv}}
=
\mathbb{E}_{(X,A)}\big[\mathcal{L}_{\text{base}}(\pi(Ie), A)\big]
+
\beta\, \hat{I}(X;Ie),
\label{eq:inv_loss}
\end{equation}
where $\pi(\cdot)$ denotes the original policy head, $\mathcal{L}_{\text{base}}$ is the standard VLA training loss, and $\hat{I}(X;Ie)$ is estimated using a neural mutual information estimator~\citep{belghazi2018mutual}. 
The first term approximates the sufficiency constraint by reducing $H(A|Ie)$, while the second term enforces information compression.
As a result, the Invariant Net captures stable correlations that better transfer under observation shifts.

\paragraph{Non-Invariant Net via Weight Subtraction.}
While the Invariant Net is encouraged to capture task-stable structure, the standard policy may retain complementary information that varies across environments.
To capture such complementary information, we construct a \emph{Non-Invariant Net} using \textbf{weight arithmetic}.
Let $W_{\text{Normal}}$ denote the weights of the standard
policy trained with $\mathcal{L}_{\text{base}}$, and
$W_{\text{Inv}}$ denote the Invariant Net weights obtained
by minimizing~\eqref{eq:inv_loss}.
We define
\begin{equation}
W_{\text{NonInv}} = W_{\text{Normal}} - W_{\text{Inv}}.
\label{eq:weight_sub}
\end{equation}
The term \emph{Non-Invariant Net} denotes this candidate
complementary branch, without assuming an exact separation
of non-invariant information.
Appendix~\ref{sec:theory} provides a local-linearization
perspective on parameter differences, while the utility
of the resulting branch is evaluated empirically.
We then compute the non-invariant embedding as
\begin{equation}
NIe = f_{W_{\text{NonInv}}}(X,L),
\end{equation}
where $f_{W_{\text{NonInv}}}$ denotes the network
parameterized by $W_{\text{NonInv}}$.

\subsection{Adaptive Mixing of Non-Invariant Information}
\label{sec:ami}

Purely invariant representations can be overly conservative, while exploiting non-invariant features may introduce spurious dependencies.
We propose Adaptive Mixing of Non-Invariant Information (\textbf{AMI}) to improve robustness to shifts in the environment-conditioned non-invariant feature distribution $\mathcal{P}(NIe \mid E)$, where $E$ denotes the environment.
AMI couples non-invariant features across samples through controlled, structure-aware mixing, thereby smoothing environment-conditioned variations in the latent space.

\paragraph{In-batch Pairing.}
For each sample $i$ in a mini-batch, we randomly sample a partner index $j \neq i$ and obtain non-invariant embeddings $(NIe_i, NIe_j)$. 
This pairing constructs a mixed embedding $\tilde{NIe}_i$ used for training, while leaving the invariant embedding $Ie_i$ unchanged.

\paragraph{Soft Mixing via Adaptive Normalization.}
We first introduce a soft mixing strategy based on Adaptive Instance Normalization (AdaIN)~\citep{huang2017arbitrary, zhou2021domaingeneralizationmixstyle}.  
We interpolate the feature statistics of two embeddings:
\begin{align}
\gamma_{\text{mix}} &= \lambda \sigma(NIe_i) + (1-\lambda)\sigma(NIe_j), \\
\beta_{\text{mix}}  &= \lambda \mu(NIe_i) + (1-\lambda)\mu(NIe_j),
\end{align}
where $\lambda \sim \text{Beta}(\alpha,\alpha)$.  
The mixed embedding is
\begin{equation}
\tilde{NIe}_i
=
\gamma_{\text{mix}}
\frac{NIe_i-\mu(NIe_i)}{\sigma(NIe_i)+\varepsilon}
+
\beta_{\text{mix}}.
\end{equation}
By interpolating feature statistics while retaining the normalized source features, soft mixing perturbs sample-specific statistics without directly interpolating the full embeddings.

\paragraph{Hard Mixing via Embedding Interpolation.}
We further consider a stronger mixing variant that directly interpolates embeddings:
\begin{equation}
\tilde{NIe}_i
=
\lambda NIe_i + (1-\lambda) NIe_j,
\end{equation}
which explicitly enforces convex interpolation between non-invariant features from different samples, thereby mitigating over-reliance on environment-specific feature modes.
In practice, soft mixing stabilizes feature statistics, while hard mixing imposes stronger cross-sample coupling; the two variants provide complementary regularization effects.

\subsection{Training Objective and Integration with VLA}
\label{sec:train}

\paragraph{Fusion and Prediction.}
Given the invariant embedding $Ie_i$ and the mixed non-invariant embedding $\tilde{NIe}_i$, we fuse them as
$
h_i = [Ie_i \,\|\, \tilde{NIe}_i],
$
and feed $h_i$ into the original policy head $\pi(\cdot)$ for action prediction. 
This design preserves the base policy interface and requires no architectural changes.

\paragraph{Stage-2 Training Objective.}
During the AMI training, we optimize the base VLA objective on the fused representation:
\begin{equation}
\mathcal{L}_{\text{MixVLA}}
=
\mathbb{E}_{(X,A)}\big[\mathcal{L}_{\text{base}}(\pi(h), A)\big],
\end{equation}
where $h$ depends on AMI through $\tilde{NIe}$. 
By exposing the policy head to the invariant structure together with regularized non-invariant cues, the model learns to reduce reliance on sample-specific correlations while retaining useful predictive information.

\section{Experiment}
\label{sec:exp}

\begin{figure*}[t]
\centering
\includegraphics[width=0.99\linewidth]{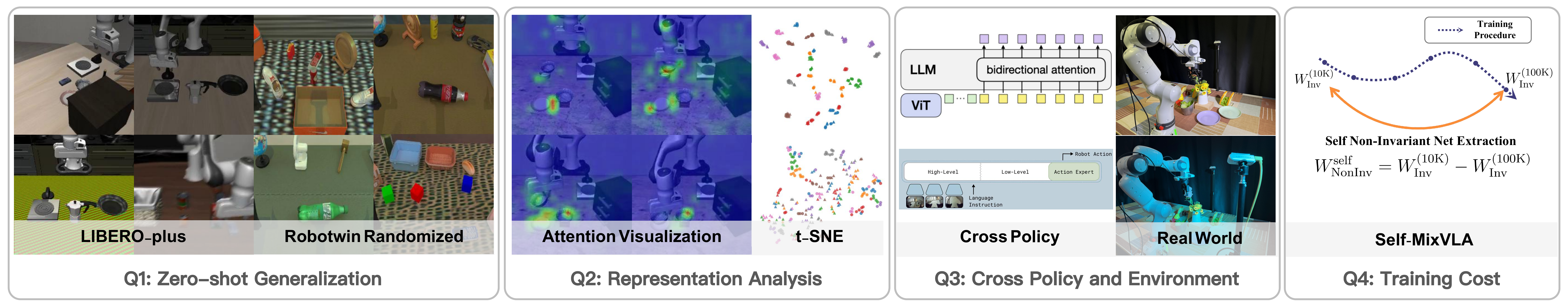}
\vspace{-5pt}
\caption{\textbf{Experimental Setup.}
Our experiments cover zero-shot generalization (Q1), representation analysis (Q2), cross-policy and real-world applicability (Q3), and training efficiency with Self-MixVLA (Q4).
}
\vspace{-5pt}
\label{fig:setup}
\end{figure*}

\begin{table}[!tp]\centering
% \vspace{-10pt}
\caption{\textbf{Performance on the LIBERO-Plus benchmark (out-of-distribution evaluation).}
}
\vspace{-8pt}
\resizebox{\linewidth}{!}{%
% \small
\begin{tabular}{lcccccccc}\toprule
\textbf{Method}&\textbf{Camera} &\textbf{Robot} &\textbf{Language} &\textbf{Light} &\textbf{Background} &\textbf{Noise} &\textbf{Layout} &\textbf{Total} \\
\midrule
OpenVLA~\citep{kim2024openvla} &0.8 &3.5 &23.0 &8.1 &34.8 &15.2 &28.5 &15.6 \\
WorldVLA~\citep{Cen2025WorldVLATA} &0.1 &27.9 &41.6 &43.7 &17.1 &10.9 &38.0 &25.0 \\
NORA~\citep{hung2025nora} &2.2 &37.0 &65.1 &45.7 &58.6 &12.8 &62.1 &39.0 \\
UniVLA~\citep{bu2025univla} &1.8 &46.2 &69.6 &69.0 &81.0 &21.2 &31.9 &42.9 \\
$\pi_0$~\citep{Black20240AV} &13.8 &6.0 &58.8 &85.0 &81.4 &79.0 &68.9 &53.6 \\
$\pi_0$-Fast~\citep{pertsch2025fastefficientactiontokenization} &\textbf{65.1} &21.6 &61.0 &73.2 &73.2 &74.4 &68.8 &61.6 \\
RIPT-VLA~\citep{tan2025interactiveposttrainingvisionlanguageactionmodels} &55.2 &31.2 &77.6 &88.4 &91.6 &73.5 &74.2 &68.4 \\
OpenVLA-OFT~\citep{kim2025fine} &56.4 &31.9 &79.5 &88.7 &93.3 &75.8 &74.2 &69.6 \\
\midrule
% \rowhighlight
\textbf{MixVLA}  
&49.6
&\textbf{63.3}
&\textbf{82.6}
&\textbf{96.1}
&\textbf{93.9}
&\textbf{82.3}
&\textbf{76.8}
&\textbf{76.2} \\
\bottomrule
\end{tabular}
}
\vspace{-10pt}
\label{tab:libero_plus_compare}
\end{table}

Our experiments assess whether regularizing, rather than discarding, non-invariant information improves zero-shot robustness to distribution shifts while retaining strong in-domain performance.
Beyond overall performance, we investigate the role of non-invariant information in generalization, the applicability of MixVLA across policy architectures and real-world settings, and the potential to reduce training requirements while preserving robustness gains
As outlined in Figure~\ref{fig:setup}, our experiments address the following four questions:

\noindent\textbf{Q1. Robust Generalization.}
Can MixVLA improve zero-shot robustness to unseen distribution
shifts while retaining strong in-domain performance?

\noindent\textbf{Q2. Representation Analysis.}
What does regularized non-invariant information contribute
beyond invariant features alone?

\noindent\textbf{Q3. General Applicability.}
Do the robustness gains extend across policy architectures
and to real-world manipulation?

\noindent\textbf{Q4. Training Cost.}
Can MixVLA retain its robustness benefits using a single
Invariant Net training run, without training a separate standard policy?

\noindent\textbf{Benchmarks.}
We evaluate MixVLA across simulation and real-world manipulation settings.
For simulation, we train on LIBERO~\citep{liu2023libero} and evaluate both in-domain performance and zero-shot robustness under distribution shifts using
LIBERO-Plus~\citep{fei2025libero}. 
We further conduct clean-to-clean (C2C) and clean-to-randomized (C2R) evaluations on RoboTwin~\citep{chen2025robotwin}, where policies are trained only with clean demonstrations to assess generalization to unseen randomized environments. 
Finally, we deploy policies on real robots and evaluate robustness.

\noindent\textbf{Implementation details.}
We implement MixVLA on two representative VLA architectures: OpenVLA-OFT~\citep{kim2025fine}, which predicts continuous action chunks through parallel decoding, and $\pi_{0.5}$~\citep{Intelligence202505AV}, which generates continuous actions through flow matching.
For the main simulation experiments, the Normal Net and the Invariant Net are each trained for 100K steps on LIBERO or clean RoboTwin demonstrations.
In Stage~2, we perform AMI training for 1K steps on LIBERO and 10K steps on RoboTwin.
Detailed training configurations are provided in the Appendix~\ref{sec:more implementation}.

% ============================================================
% Q1: Does robustness improve without losing in-domain ability?
% ============================================================

\subsection{Robustness to Unseen Distribution Shifts}
\label{sec:exp_robustness}

\begin{table}[t]
\centering

\caption{\textbf{Comparison on the LIBERO benchmark (in-domain setting). }
}
\vspace{-8pt}
\small
\renewcommand{\arraystretch}{1.0}
\begin{tabularx}{\linewidth}{
    l
    >{\centering\arraybackslash}X
    >{\centering\arraybackslash}X
    >{\centering\arraybackslash}X
    >{\centering\arraybackslash}X
    >{\centering\arraybackslash}X
}
\toprule
\textbf{Model} & \textbf{Spatial} & \textbf{Object} & \textbf{Goal} & \textbf{Long} & \textbf{Avg} \\
\midrule
OpenVLA~\citep{kim2024openvla}     & 84.7 & 88.4 & 79.2 & 53.7 & 76.5 \\
WorldVLA~\citep{Cen2025WorldVLATA}     & 85.6 & 89.0 & 82.6 & 59.0 & 79.1 \\
NORA~\citep{hung2025nora}     & 92.2 & 95.4 & 89.4 & 74.6 & 87.9 \\
UniVLA~\citep{bu2025univla}     & 96.5 & 96.8 & 95.6 & 92.0 & 95.2 \\
$\pi_0$~\citep{Black20240AV}     & 90.0 & 86.0 & 95.0 & 73.0 & 86.0 \\
$\pi_0$-Fast~\citep{pertsch2025fastefficientactiontokenization}     & 96.4 & 96.8 & 88.6 & 60.2 & 85.5 \\
OpenVLA-OFT~\citep{kim2025fine}     & \csecond{97.6} & \cfirst{98.4} & \cfirst{97.9} & \cfirst{94.5} & \cfirst{97.1} \\
\midrule
\textbf{MixVLA} & \cfirst{98.4} & \csecond{97.2} & \csecond{97.4} & \csecond{93.6} & \csecond{96.7} \\
\bottomrule
\end{tabularx}
\label{tab:libero_compare}
% \vspace{-8pt}
\end{table}

\noindent\textbf{Robustness from LIBERO to LIBERO-Plus.}
We build MixVLA on OpenVLA-OFT and train it on LIBERO, evaluating on LIBERO and LIBERO-Plus.
As shown in Table~\ref{tab:libero_plus_compare}, MixVLA achieves the highest overall success rate of 76.2\% among the compared methods, surpassing OpenVLA-OFT by 6.6\%.
However, its lower performance under camera perturbations suggests that non-invariant feature mixing alone may be insufficient to address viewpoint-induced geometric shifts.
On LIBERO, MixVLA retains strong in-domain performance, achieving an average success rate of 96.7\% compared with 97.1\% for OpenVLA-OFT (Table~\ref{tab:libero_compare}),
with a decrease of only 0.4\%.

\begin{table}[t]
    \centering
    \vspace{-10pt}
    \caption{
    \textbf{Clean-to-clean and clean-to-randomized evaluation on RoboTwin 2.0.}
    }
    \label{tab:robotwin_c2r}
    \vspace{-8pt}

    \begin{tabularx}{0.99\linewidth}{
        l
        >{\centering\arraybackslash}X
        >{\centering\arraybackslash}X
        >{\centering\arraybackslash}X
    }
        \toprule
        \textbf{Method}
        & \textbf{C2C}
        & \textbf{C2R}
        & \textbf{Avg.} \\
        \midrule
        Fast-WAM~\citep{yuan2026fastwam}
            & \csecond{77.8} & 1.9 & 39.9 \\
        X-VLA~\citep{zheng2026x}
            & 68.0 & 20.9 & 44.5 \\
        X-WAM~\citep{guo2026xwam}
            & 70.0 & 25.8 & 47.9 \\
        Spatial Forcing~\citep{li2026spatial}
            & 77.2 & 26.7 & 52.0 \\
        $\pi_{0.5}$~\citep{Intelligence202505AV}
            & 70.7 & \csecond{46.0} & 58.4 \\
        4D-WAM~\citep{yang20264d}
            & \cfirst{\textbf{81.5}}
            & 41.8
            & \csecond{61.6} \\
        \midrule
        \textbf{MixVLA}
            & 75.6
            & \cfirst{\textbf{62.4}}
            & \cfirst{\textbf{69.0}} \\
        \bottomrule
    \end{tabularx}
    \vspace{-10pt}
\end{table}

\noindent\textbf{Clean-to-randomized generalization on RoboTwin.}
We build MixVLA on $\pi_{0.5}$ and train it exclusively on clean demonstrations from RoboTwin 2.0.
As shown in Table~\ref{tab:robotwin_c2r}, MixVLA improves C2R success from 46.0\% to 62.4\% (+16.4\%) over $\pi_{0.5}$, while also increasing C2C success from 70.7\% to 75.6\% (+4.9\%).

Taken together, these results support a positive answer to \textbf{Q1:} MixVLA improves robustness to a range of unseen distribution shifts while retaining strong in-domain performance.

% ============================================================
% Q2: What does non-invariant information contribute?
% ============================================================

\subsection{Understanding the Role of Non-Invariant Information}
\label{sec:exp_noninvariant}

\begin{table}[t]
\centering
\vspace{-2pt}
\caption{\textbf{Ablation study of IB and AMI on LIBERO-Plus
and RoboTwin 2.0.}
Experiments use OpenVLA-OFT on LIBERO-Plus and $\pi_{0.5}$
on RoboTwin 2.0.
}
\vspace{-8pt}
\label{tab:ablation}
\small
\setlength{\tabcolsep}{4pt}
\renewcommand{\arraystretch}{1.1}
\resizebox{\linewidth}{!}{%
\begin{tabular}{cc cccccccc ccc}
\toprule
\multirow{2}{*}{\textbf{IB}}
& \multirow{2}{*}{\textbf{AMI}}
& \multicolumn{8}{c}{\textbf{LIBERO-Plus}}
& \multicolumn{3}{c}{\textbf{RoboTwin}} \\
\cmidrule(lr){3-10}
\cmidrule(lr){11-13}
&
& \textbf{Camera}
& \textbf{Robot}
& \textbf{Language}
& \textbf{Lighting}
& \textbf{Background}
& \textbf{Noise}
& \textbf{Layout}
& \textbf{Total}
& \textbf{C2C}
& \textbf{C2R}
& \textbf{Avg.} \\
\midrule
&
& \textbf{51.2}
& 30.7
& 74.2
& 87.8
& 92.6
& 76.0
& 73.7
& 67.5
& 73.8
& 53.8
& 63.8 \\
\checkmark &
& 48.8
& 58.6
& 72.7
& 91.3
& 93.0
& 78.8
& 74.1
& 72.2
& \textbf{75.8}
& 59.6
& 67.7 \\
\checkmark & \checkmark
& 49.6
& \textbf{63.3}
& \textbf{82.6}
& \textbf{96.1}
& \textbf{93.9}
& \textbf{82.3}
& \textbf{76.8}
& \textbf{76.2}
& 75.6
& \textbf{62.4}
& \textbf{69.0} \\
\bottomrule
\end{tabular}%
}
\vspace{-5pt}
\end{table}

\noindent\textbf{Beyond invariant representations.}
We compare the Normal Net, Invariant Net, and full MixVLA, using OpenVLA-OFT on LIBERO-Plus and $\pi_{0.5}$ on RoboTwin (Table~\ref{tab:ablation}).
IB-only (Invariant Net) improves overall OOD performance on both benchmarks, but reduces robustness under camera and language perturbations on LIBERO-Plus relative to the standard policy.
This uneven behavior highlights the limitations of relying solely on invariant representations to handle diverse distribution shifts.
Complementing invariant representations with AMI-regularized non-invariant features yields further gains of 4.0\% on LIBERO-Plus and 2.8\% on RoboTwin C2R over IB-only.
MixVLA outperforms Invariant Net across all seven LIBERO-Plus perturbation categories while largely preserving clean-environment performance on RoboTwin.
These results support our central premise that non-invariant information need not be discarded: when regularized, it can complement invariant features and improve OOD generalization.
Additional analyses of mixing strategies are provided in the Appendix~\ref{sec:more experiments}.

\begin{figure}[t]
\centering

\begin{subfigure}[t]{0.62\linewidth}
    \vspace{0pt}
    \centering
    \includegraphics[width=\linewidth]{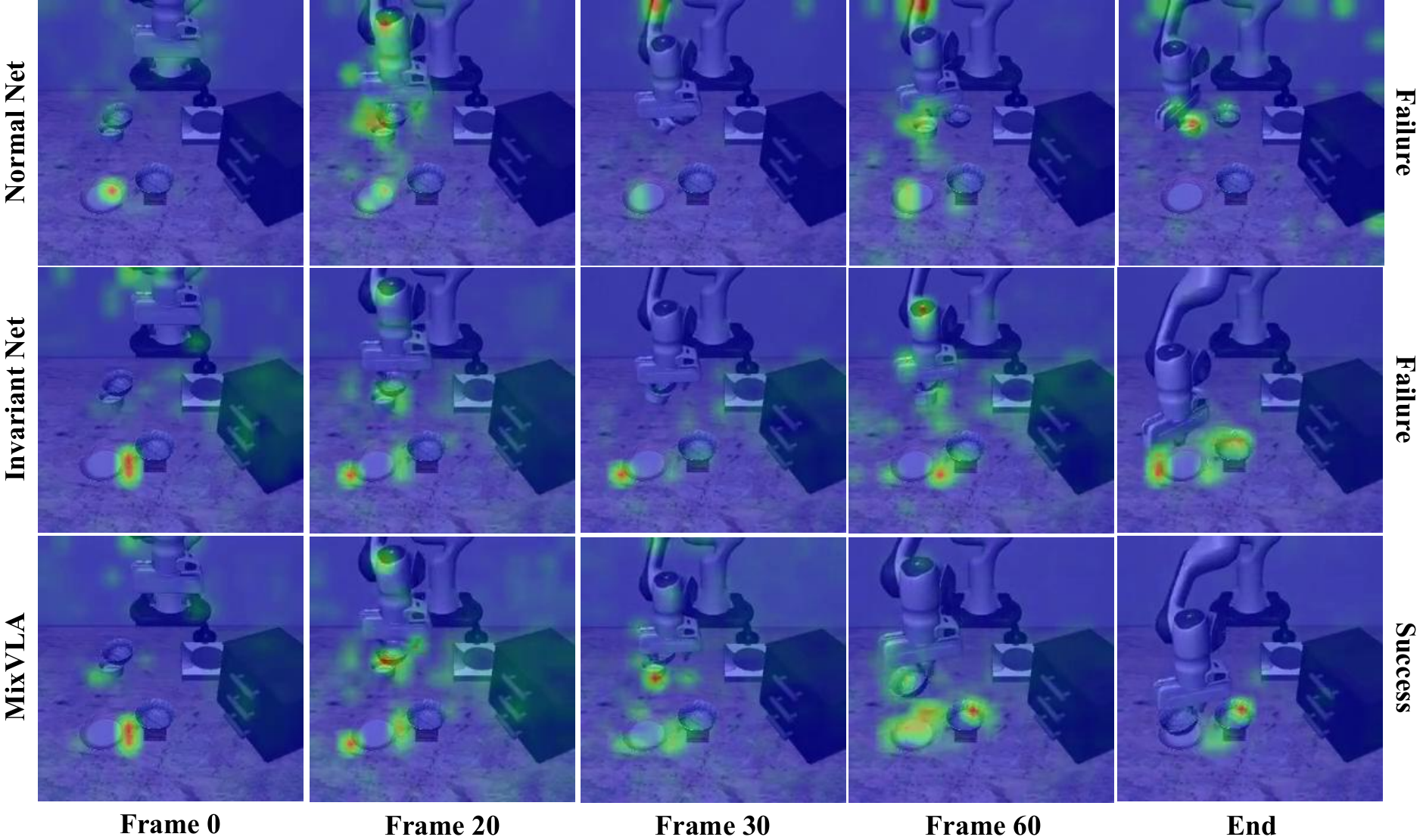}
    \caption{Grad-CAM on LIBERO-Plus.}
    \label{fig:att_vis}
\end{subfigure}
\hfill
\begin{subfigure}[t]{0.35\linewidth}
    \vspace{0pt}
    \centering
    % Prefer a 2-by-2 panel layout for the t-SNE figure.
    \includegraphics[width=\linewidth]{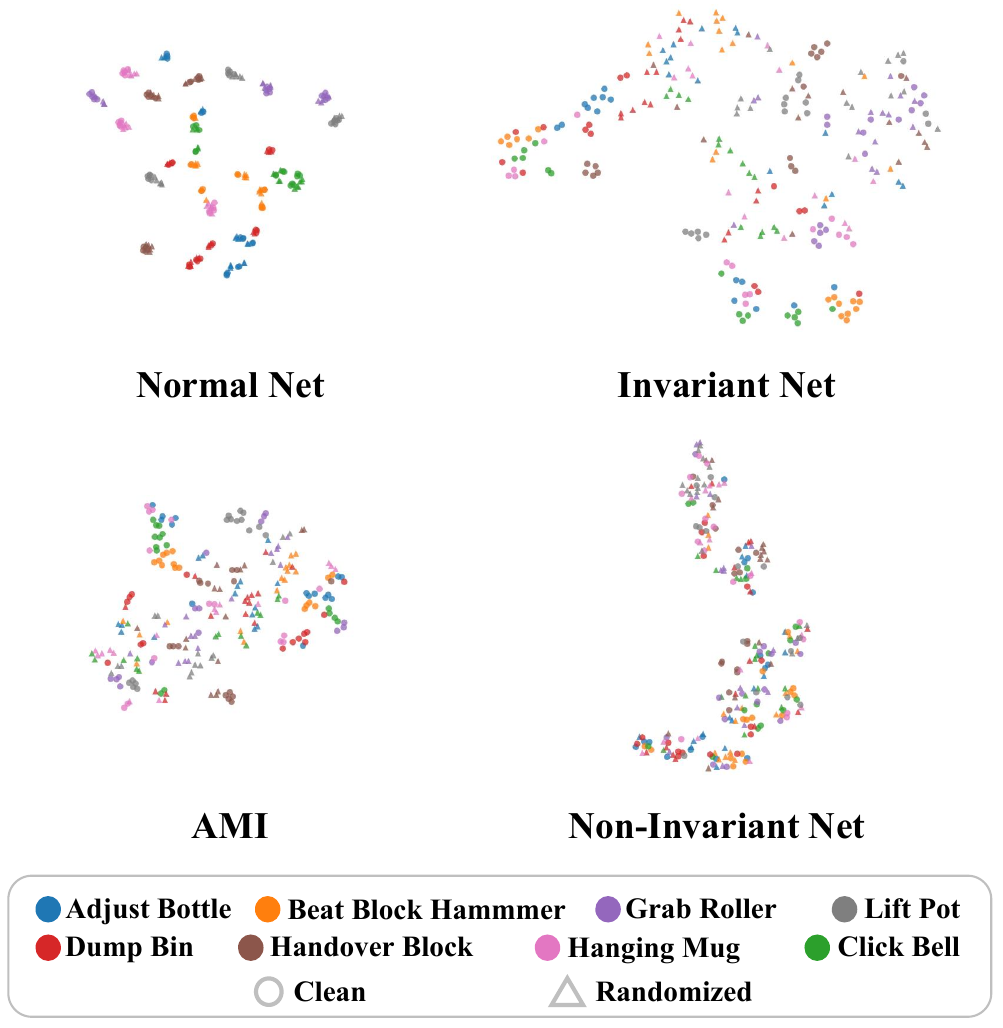}
    \caption{t-SNE on RoboTwin.}
    \label{fig:tsne}
\end{subfigure}
\vspace{-5pt}
\caption{\textbf{Qualitative analysis under distribution shifts.}
(a) Grad-CAM visualizations along each policy's own rollout.
(b) t-SNE projections of mean-pooled token features from
eight tasks.
Colors indicate tasks; circles and triangles denote clean and randomized observations, respectively.
% Each representation is projected independently, so coordinates are not directly comparable across panels.
}
\vspace{-15pt}
\label{fig:qualitative}
\end{figure}

\noindent\textbf{Visual attribution under distribution shifts.}
Figure~\ref{fig:att_vis} presents Grad-CAM visualizations along the respective rollouts of the Normal Net, Invariant Net, and MixVLA on a LIBERO-Plus task.
The Invariant Net exhibits localized responses around manipulation-related regions, whereas MixVLA highlights these regions together with broader nearby context at intermediate steps.
Both the Normal Net and the Invariant Net fail to complete the task, while MixVLA succeeds.
Figure~\ref{fig:tsne} further visualizes mean-pooled token features from eight RoboTwin tasks under clean and randomized conditions.
The Normal Net forms predominantly task-homogeneous local clusters, whereas the Invariant Net projection exhibits a different organization with some regions still dominated by either clean or randomized samples.
MixVLA retains task-associated neighborhoods alongside local clean--randomized interleaving, while the Non-Invariant Net exhibits substantial cross-task overlap.
These projections illustrate changes in feature organization rather than establish complete domain alignment.

Together with the ablations in Table~\ref{tab:ablation}, these qualitative observations support our answer to \textbf{Q2}: regularized non-invariant information can complement invariant features and improve OOD generalization.

% ============================================================
% Q3: Do the gains extend across policies and to real robots?
% ============================================================

\subsection{Generality across Policies and Real-World Deployment}
\label{sec:exp_generality}

\paragraph{Applicability across policy architectures.}
As shown in Tables~\ref{tab:libero_plus_compare} and~\ref{tab:robotwin_c2r}, MixVLA improves robustness over the corresponding baselines for both OpenVLA-OFT and $\pi_{0.5}$.
These gains are obtained using the same two-stage training framework, despite their different action-generation mechanisms: parallel action decoding in OpenVLA-OFT and flow matching in $\pi_{0.5}$.
These results support the model-agnostic design of MixVLA, demonstrating its applicability beyond a single policy architecture.

\begin{wrapfigure}{l}{0.48\textwidth}
\centering
% \vspace{-10pt}
\includegraphics[width=\linewidth]{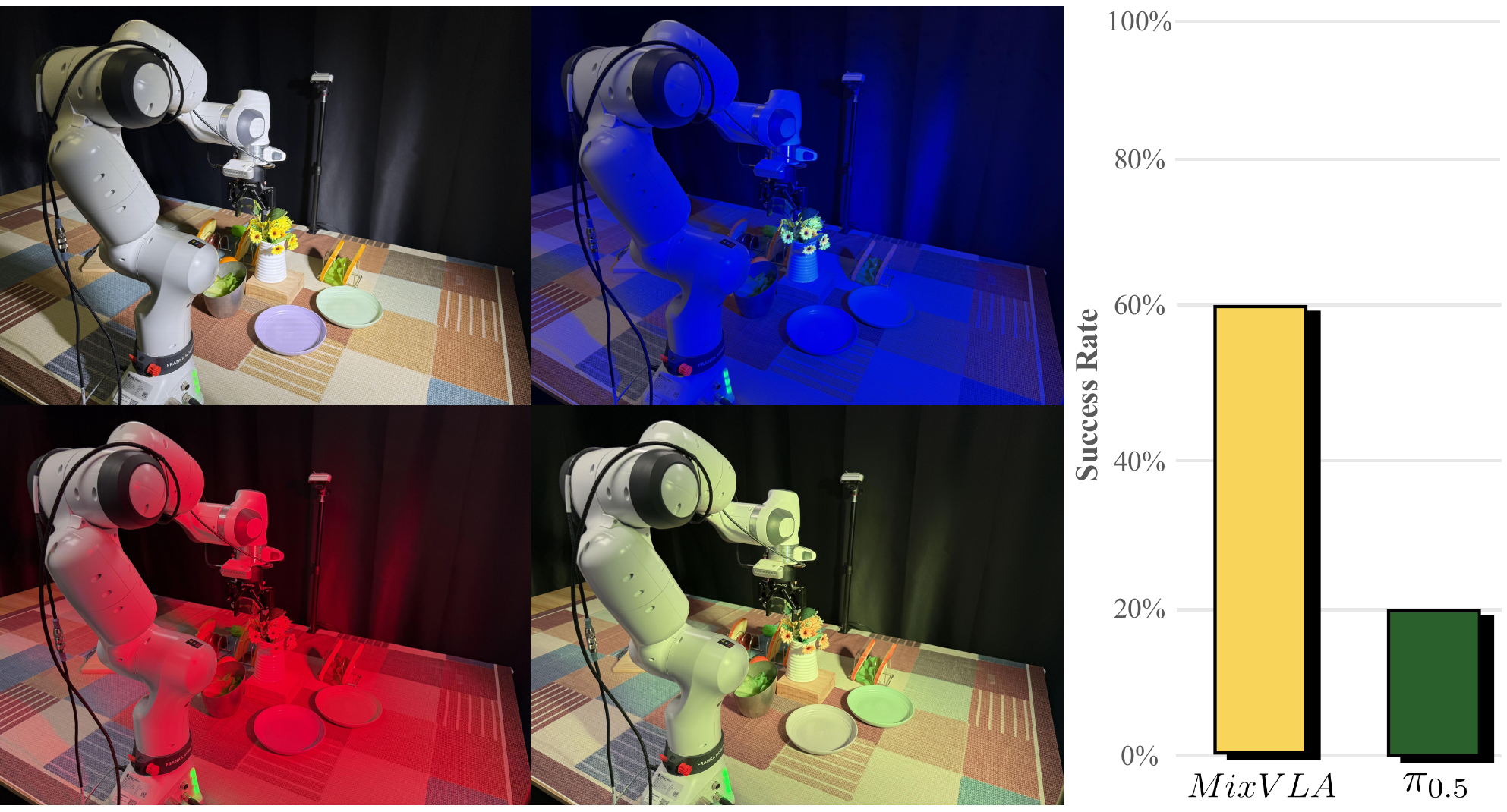}
\vspace{-10pt}
\caption{\textbf{Real-world robustness to lighting changes.}
}
\vspace{-10pt}
\label{fig:realworld}
\end{wrapfigure}

\paragraph{Robustness in real-world manipulation.}
We deploy $\pi_{0.5}$ and its MixVLA counterpart on a Franka robot.
We collect 80 demonstrations for a bread-placement task with two instruction variants, requiring the robot to pick up the sixth slice of bread and place it on either the near or far plate, as specified by the instruction.
We evaluate each policy over 10 trials under lighting perturbations, counting a trial as successful when the specified slice is placed on the instructed plate.
As shown in Figure~\ref{fig:realworld}, MixVLA achieves a success rate of 60\%, compared with 20\% for $\pi_{0.5}$. See Appendix~\ref{sec:more experiments} for videos of real-world executions.

These results complement the cross-architecture simulation findings and provide preliminary evidence that MixVLA's robustness benefits extend to real-world manipulation, supporting a positive answer to \textbf{Q3}.

% ============================================================
% Q4: Is a separately trained standard policy necessary?
% ============================================================

\subsection{Robustness from a Single Training Run}
\label{sec:exp_single_run}

\paragraph{Reusing intermediate IB checkpoints.}
\begin{wraptable}{r}{0.48\textwidth}
% \vspace{-20pt}
\centering
\caption{\textbf{Single-run variant on RoboTwin 2.0.}
}
\vspace{-5pt}
\label{tab:single_run}
\small
\setlength{\tabcolsep}{3pt}
\renewcommand{\arraystretch}{1.1}
\begin{tabularx}{\linewidth}{
    @{}l
    >{\centering\arraybackslash}X
    >{\centering\arraybackslash}X
    >{\centering\arraybackslash}X@{}
}
\toprule
\textbf{Method}
& \textbf{C2C}
& \textbf{C2R}
& \textbf{Avg.} \\
\midrule
Self-MixVLA
& 73.0
& 60.8
& 66.9 \\
MixVLA
& \textbf{75.6}
& \textbf{62.4}
& \textbf{69.0} \\
\bottomrule
\end{tabularx}
% \vspace{-10pt}
\end{wraptable}
We explore whether intermediate checkpoints from a single Invariant Net training run can eliminate the need to train a separate Normal Net.
Our intuition is that Invariant Net training progressively favors invariant information, and differences between early and late checkpoints may provide a proxy for complementary non-invariant information.
Motivated by this hypothesis, we introduce \textbf{Self-MixVLA}, which constructs a \emph{Self Non-Invariant Net} from the 10K- and 100K-step checkpoints of the same IB training run:
\begin{equation}
W_{\text{NonInv}}^{\text{self}}
=
W_{\text{Inv}}^{(10\mathrm{K})}
-
W_{\text{Inv}}^{(100\mathrm{K})}.
\label{eq:self_weight_sub}
\end{equation}
The resulting branch is paired with the 100K-step Invariant Net for subsequent AMI training, avoiding a separate Normal Net training run.

These results support a positive answer to \textbf{Q4}: Self-MixVLA reduces training requirements
by eliminating the separate Normal Net training run, while retaining robustness benefits with a modest performance reduction relative to full MixVLA (Table~\ref{tab:single_run}).
We regard this variant as a preliminary exploration of reusing a single IB training trajectory, with further opportunities to investigate checkpoint selection and extrapolation strategies for recovering complementary information.

\section{Conclusion}

In this work, we presented \textbf{MixVLA}, a model-agnostic
framework that improves VLA generalization by regularizing
rather than discarding non-invariant information.
MixVLA learns invariant representations through an information
bottleneck objective and derives a complementary branch through
weight arithmetic.
\textbf{AMI} regularizes
non-invariant features through stochastic mixing before fusing
them with invariant representations for action prediction.
Experiments demonstrate improved overall zero-shot robustness
while retaining strong in-domain performance.
Our findings suggest that non-invariant information can provide
useful predictive cues when appropriately regularized, offering
a practical direction for generalizable embodied policies
without additional OOD training data.

\bibliography{iclr2027_conference}
\bibliographystyle{iclr2027_conference}

\appendix
\maketitlesupplementary
\begin{abstract}
    \noindent This supplementary material expands upon our main paper by providing additional theoretical analysis, implementation details, experimental results, and discussions to further support the proposed method \ours. \\
    \noindent $\triangleright$ \textbf{Sec.~\ref{sec:theory}}: We provide theoretical analyses of MixVLA, including the information-compression effect of IB, the statistical properties of adaptive mixing, and an illustrative regression analysis that characterizes when regularizing complementary information can improve OOD robustness. \\
    \noindent $\triangleright$ \textbf{Sec.~\ref{sec:more implementation}}: We provide additional implementation details, including model architectures, optimization settings, and training configurations. \\
    \noindent $\triangleright$ \textbf{Sec.~\ref{sec:more experiments}}: We report additional experiments and qualitative results, including ablations on mixing strategies, more visualizations on LIBERO, LIBERO-Plus, RoboTwin, and real-world demonstration videos. \\
    \noindent $\triangleright$ \textbf{Sec.~\ref{sec:discussion}}: We discuss the limitations of the current framework and outline several possible directions for future research.
\end{abstract}

\section{Theoretical Analysis}
\label{sec:theory}

This section analyzes the information-compression effect
of the Information Bottleneck (IB), the statistical
properties of adaptive mixing, and a local-linearization
interpretation of MixVLA.
We also present an illustrative regression model in which
mixing improves OOD prediction under explicit conditions.
These analyses characterize aspects of the proposed
framework rather than establish unconditional guarantees
for nonlinear VLA policies.

Throughout this section, $I(\cdot;\cdot)$ and
$I(\cdot;\cdot\mid\cdot)$ denote mutual information and
conditional mutual information, respectively.
For brevity, we write $T=Ie$ and $N=NIe$.
Importantly, the learned complementary representation $N$
is not assumed to contain only nuisance information.
All information quantities and moments used below are
assumed finite.
Language instructions are treated as fixed context and
omitted from the notation.

\subsection{Invariant Representation via Information Bottleneck}

The IB objective balances input compression against
predictive information retention
\citep{tishby2000informationbottleneckmethod}.
Related analyses connect representation minimality with
insensitivity to nuisance factors under suitable assumptions
\citep{achille2018emergenceinvariancedisentanglementdeep}.
Here, we characterize the conditional nuisance information
controlled by the compression objective.

Assume the observation is generated as
\begin{equation}
X = S(Z_{\mathrm{inv}}, Z_{\mathrm{ninv}}),
\end{equation}
where $Z_{\mathrm{inv}}$ contains task-relevant latent
factors and $Z_{\mathrm{ninv}}$ contains
environment-dependent factors.
Assume
\begin{equation}
A \perp X \mid Z_{\mathrm{inv}},
\end{equation}
and let $T$ be sampled from an encoder $p_\phi(T\mid X)$,
whose sampling randomness is independent of the other
variables conditional on $X$.

We consider the constrained IB objective
\begin{equation}
\min_{p_\phi(T\mid X)} I(X;T)
\quad \text{s.t.} \quad
I(T;A) \geq \kappa,
\label{eq:supp_ib_objective}
\end{equation}
where $\kappa>0$ is chosen so that the constraint is feasible.

\noindent\textbf{Proposition A.1
(An upper bound on conditional nuisance information).}
Under the assumptions above,
\begin{equation}
\begin{aligned}
I(T;Z_{\mathrm{ninv}}\mid Z_{\mathrm{inv}})
&= I(X;T)-I(T;Z_{\mathrm{inv}}) \\
&\leq I(X;T)-I(T;A).
\end{aligned}
\label{eq:supp_ib_nuisance_bound}
\end{equation}
Consequently, any feasible encoder in
Eq.~\eqref{eq:supp_ib_objective} satisfies
\begin{equation}
I(T;Z_{\mathrm{ninv}}\mid Z_{\mathrm{inv}})
\leq I(X;T)-\kappa.
\end{equation}

\noindent\textbf{Proof.}
Since $X$ is a deterministic function of
$(Z_{\mathrm{inv}},Z_{\mathrm{ninv}})$ and the encoder
depends on these factors only through $X$,
\begin{equation}
I(X;T)
=
I(Z_{\mathrm{inv}},Z_{\mathrm{ninv}};T).
\end{equation}
Applying the chain rule gives
\begin{equation}
I(X;T)
=
I(Z_{\mathrm{inv}};T)
+
I(Z_{\mathrm{ninv}};T\mid Z_{\mathrm{inv}}).
\end{equation}
Moreover, the action-sufficiency assumption and the
encoder construction imply the Markov chain
$A \leftrightarrow Z_{\mathrm{inv}} \leftrightarrow T$.
The data processing inequality therefore yields
\begin{equation}
I(T;A)\leq I(T;Z_{\mathrm{inv}}).
\end{equation}
Combining these relations proves the result.
\hfill$\square$

\noindent\textbf{Interpretation.}
At a prescribed level of predictive information,
reducing $I(X;T)$ tightens an upper bound on the
information retained about environment-dependent factors
beyond the task-relevant latent factors.
This provides an information-theoretic motivation for
IB regularization, but does not imply that a practical
encoder eliminates all nuisance dependence or generalizes
to arbitrary unseen environments.

The ideal latent factors $Z_{\mathrm{inv}}$ should also
be distinguished from the learned representation $T$.
Finite-capacity encoders and practical IB optimization
need not make $T$ sufficient for action prediction.
A complementary representation can therefore retain
useful information that is missing from $T$, motivating
regularization rather than unconditional removal of
the complementary branch.

\subsection{Generalization Benefit of Adaptive Mixing}

We first characterize the moments of hard mixing and
then identify conditions under which it improves
prediction in an illustrative regression model.

\noindent\textbf{Moments of hard mixing.}
Let $N_i$ and $N_j$ be independent samples from the same
source feature distribution, with mean $\mu$ and
covariance $\Sigma$.
Let $\lambda\sim\mathrm{Beta}(\alpha,\alpha)$,
$\alpha>0$, be independent of the samples, and define
\begin{equation}
\widetilde{N}_i
=
\lambda N_i+(1-\lambda)N_j.
\label{eq:supp_hard_mixing}
\end{equation}
Write
\begin{equation}
m_\alpha=\mathbb{E}[\lambda]=\frac{1}{2},
\qquad
q_\alpha
=
\mathbb{E}\!\left[\lambda^2+(1-\lambda)^2\right]
=
\frac{\alpha+1}{2\alpha+1}.
\label{eq:supp_mixing_moments}
\end{equation}
Then
\begin{equation}
\mathbb{E}[\widetilde{N}_i]=\mu,
\qquad
\operatorname{Cov}(\widetilde{N}_i)
=
q_\alpha\Sigma.
\label{eq:supp_mixed_covariance}
\end{equation}

To obtain Eq.~\eqref{eq:supp_mixed_covariance}, note that
the conditional mean given $\lambda$ is $\mu$, while
independence gives
\begin{equation}
\operatorname{Cov}(\widetilde{N}_i\mid\lambda)
=
\left[\lambda^2+(1-\lambda)^2\right]\Sigma.
\end{equation}
The result follows from the law of total covariance.
Since $1/2<q_\alpha<1$, mixing contracts the
across-sample covariance in this population model.

This contraction alone does not establish improved
prediction: useful signals can also be attenuated.
The following model makes the resulting trade-off explicit.

\noindent\textbf{An illustrative regression model.}
Consider a scalar action
\begin{equation}
A=T+C,
\end{equation}
where $T$ is the contribution captured by the invariant
representation and $C$ is a complementary predictive
signal not captured by $T$.
Assume the complementary feature is
\begin{equation}
N_{\mathrm{src}}=C+\epsilon_{\mathrm{src}},
\qquad
N_{\mathrm{tgt}}=C+\epsilon_{\mathrm{tgt}}.
\end{equation}
The variables $T$, $C$, and the noise terms are
mutually independent and centered, with
\begin{equation}
\operatorname{Var}(C)=v>0,
\qquad
\operatorname{Var}(\epsilon_{\mathrm{src}})
=\sigma_{\mathrm{src}}^2,
\qquad
\operatorname{Var}(\epsilon_{\mathrm{tgt}})
=\sigma_{\mathrm{tgt}}^2.
\end{equation}
The distributions of $T$ and $C$ remain unchanged
between source and target environments.

We restrict attention to predictors
\begin{equation}
f_w(T,N)=T+wN,
\end{equation}
so that invariant-only prediction corresponds to $w=0$.
ERM minimizes population squared error using
$N_{\mathrm{src}}$.
AMI instead minimizes squared error using
$\widetilde{N}_i$ from Eq.~\eqref{eq:supp_hard_mixing},
while retaining the original target $A_i$.
The partner feature is sampled independently, and
unmixed features are used at evaluation.

Define
\begin{equation}
V_{\mathrm{src}}=v+\sigma_{\mathrm{src}}^2,
\qquad
V_{\mathrm{tgt}}=v+\sigma_{\mathrm{tgt}}^2,
\qquad
r_\alpha=\frac{m_\alpha}{q_\alpha}
=\frac{2\alpha+1}{2\alpha+2}.
\end{equation}

\noindent\textbf{Theorem A.2
(Conditional OOD benefit in the regression model).}
Under the assumptions above, the population-optimal
coefficients are
\begin{equation}
w_{\mathrm{ERM}}=\frac{v}{V_{\mathrm{src}}},
\qquad
w_{\mathrm{AMI}}=r_\alpha w_{\mathrm{ERM}}.
\label{eq:supp_optimal_coefficients}
\end{equation}
If
\begin{equation}
\frac{2V_{\mathrm{src}}}{1+r_\alpha}
<
V_{\mathrm{tgt}}
<
\frac{2V_{\mathrm{src}}}{r_\alpha},
\label{eq:supp_ood_condition}
\end{equation}
then AMI has strictly lower target squared error
than both ERM and invariant-only prediction.

\noindent\textbf{Proof.}
For the source distribution,
$\mathbb{E}[N_{\mathrm{src}}C]=v$ and
$\mathbb{E}[N_{\mathrm{src}}^2]=V_{\mathrm{src}}$.
Thus,
\begin{equation}
w_{\mathrm{ERM}}
=
\frac{\mathbb{E}[N_{\mathrm{src}}C]}
{\mathbb{E}[N_{\mathrm{src}}^2]}
=
\frac{v}{V_{\mathrm{src}}}.
\end{equation}
For mixed features, independence of the partner sample
gives
\begin{equation}
\mathbb{E}[\widetilde{N}_i C_i]=m_\alpha v,
\qquad
\mathbb{E}[\widetilde{N}_i^2]=q_\alpha V_{\mathrm{src}}.
\end{equation}
This proves Eq.~\eqref{eq:supp_optimal_coefficients}.

At evaluation, the target risk is
\begin{equation}
R_{\mathrm{tgt}}(w)
=
\mathbb{E}\!\left[(C-wN_{\mathrm{tgt}})^2\right]
=
v-2vw+V_{\mathrm{tgt}}w^2.
\end{equation}
Let $w_0=w_{\mathrm{ERM}}$ and $r=r_\alpha$.
The risk differences satisfy
\begin{align}
R_{\mathrm{tgt}}(rw_0)-R_{\mathrm{tgt}}(w_0)
&=
(1-r)w_0
\left[2v-(1+r)w_0V_{\mathrm{tgt}}\right],
\\
R_{\mathrm{tgt}}(rw_0)-R_{\mathrm{tgt}}(0)
&=
rw_0
\left[rw_0V_{\mathrm{tgt}}-2v\right].
\end{align}
Since $0<r<1$ and $w_0=v/V_{\mathrm{src}}$,
both differences are negative under
Eq.~\eqref{eq:supp_ood_condition}.
\hfill$\square$

\noindent\textbf{Interpretation and scope.}
In this model, AMI shrinks reliance on the complementary
feature without removing it.
When target noise increases within the stated range,
this balances the cost of using a noisy signal against
the cost of discarding a useful signal.
The result establishes a setting in which regularized
complementary information is beneficial, rather than
a universal ranking of AMI, ERM, and invariant-only
prediction.

The model assumes independent pairing, population
optimization, a linear predictor, and a particular
noise-variance shift.
It does not establish that real VLA features satisfy
these assumptions, and the benefit for the proposed
policies is evaluated empirically.

\subsection{Special Cases of AMI}

The two mixing strategies operate on different quantities.
Hard mixing interpolates complete embeddings, whereas
soft mixing interpolates their feature statistics.

\noindent\textbf{Hard mixing.}
Equation~\eqref{eq:supp_hard_mixing} places the mixed
representation on the line segment joining the two
sampled embeddings.
Under independent sampling, its covariance satisfies
Eq.~\eqref{eq:supp_mixed_covariance}.
This result concerns the source feature distribution;
it does not imply averaging over unseen environments
or preservation of every action-relevant feature.
The OOD benefit in Theorem~A.2 applies only under the
additional regression assumptions stated there.

\noindent\textbf{Soft mixing via adaptive normalization.}
Following the feature-statistic formulation of adaptive
normalization and statistic mixing
\citep{huang2017arbitrary,zhou2021domaingeneralizationmixstyle},
let
\begin{equation}
\mu_i=\mu(N_i),
\qquad
\sigma_i=\sigma(N_i),
\qquad
R_i=\frac{N_i-\mu_i}{\sigma_i+\varepsilon},
\end{equation}
where the statistics are computed along the specified
aggregation dimension and arithmetic is elementwise.
Soft mixing forms
\begin{align}
\mu_{\mathrm{mix}}
&=\lambda\mu_i+(1-\lambda)\mu_j,\\
\sigma_{\mathrm{mix}}
&=\lambda\sigma_i+(1-\lambda)\sigma_j,
\end{align}
and reconstructs
\begin{equation}
\widetilde{N}^{\mathrm{soft}}_i
=
\sigma_{\mathrm{mix}}\odot R_i
+
\mu_{\mathrm{mix}}.
\label{eq:supp_soft_mixing}
\end{equation}

Define the statistic vector
\begin{equation}
s_i=
\begin{bmatrix}
\mu_i\\
\sigma_i
\end{bmatrix}.
\end{equation}
Its mixed counterpart is
$s_{\mathrm{mix}}=\lambda s_i+(1-\lambda)s_j$.
If $s_i$ and $s_j$ are independent samples with mean
$\mu_s$ and covariance $\Sigma_s$, the same moment
calculation gives
\begin{equation}
\mathbb{E}[s_{\mathrm{mix}}]=\mu_s,
\qquad
\operatorname{Cov}(s_{\mathrm{mix}})
=
q_\alpha\Sigma_s.
\end{equation}

Thus, soft mixing contracts the across-sample covariance
of the statistic vectors in this idealized setting,
while retaining the normalized source pattern $R_i$
in the reconstruction.
This does not imply that
$\operatorname{Cov}(\widetilde{N}^{\mathrm{soft}}_i)$
equals $q_\alpha\operatorname{Cov}(N_i)$, because
$R_i$ can be statistically dependent on the source
statistics.
Preservation of the normalized pattern also does not,
by itself, guarantee preservation of task semantics.

Soft and hard mixing therefore provide distinct forms
of feature regularization.
Their effectiveness for action prediction depends on
the structure of the learned representations and is
examined in our experiments.

\section{Implementation Details}
\label{sec:more implementation}

\paragraph{Backbones and fine-tuning.}
We implement MixVLA on OpenVLA-OFT for LIBERO/LIBERO-Plus and on
$\pi_{0.5}$ for RoboTwin and real-world manipulation.
For OpenVLA-OFT, we adapt the pretrained 7B model using LoRA with rank 32, scaling parameter 16, and zero dropout, targeting linear layers.
The base weights remain frozen, while the adapters, continuous action head, and proprioceptive projector are trainable.
The model receives an external camera image, a wrist camera image, the language instruction, and proprioceptive state. 
It predicts chunks of eight 7-dimensional actions using an $L_1$ regression objective.
Training uses image augmentation and quantile-based action and proprioception normalization.

For $\pi_{0.5}$, we fully fine-tune the policy, including the visual encoder, with the flow-matching objective. 
Inputs comprise three RGB views resized to $224\times224$, the instruction, and robot state;
The action horizon is 50. 
Normalization statistics are computed from the corresponding training split. 
We use bfloat16 backbone computation, gradient checkpointing, and a training seed of 42.
The OpenVLA experiments use eight NVIDIA A100 GPUs with 80 GB memory each. Training is distributed across eight GPUs. 
The global batch size is 64 for $\pi_{0.5}$; the OpenVLA configuration uses 64 for BC/IB and 48 for AMI, without gradient accumulation.

\paragraph{Optimization and schedules.}
Normal Net and Invariant Net are independently initialized from the pretrained backbone.
AMI initializes its main branch from the selected Invariant Net checkpoint and
uses a newly initialized fusion module and optimizer.
OpenVLA uses AdamW with $(\beta_1,\beta_2)=(0.9,0.999)$,
$\epsilon=10^{-8}$, and weight decay $10^{-2}$.
Its Normal Net and Invariant Net schedules have no warmup and reduce the learning
rate by a factor of ten at step 100k. 
The AMI learning rate remains constant during its short training stage.
For $\pi_{0.5}$, AdamW uses $(\beta_1,\beta_2)=(0.9,0.95)$,
$\epsilon=10^{-8}$, weight decay $10^{-10}$, and gradient-norm
clipping at 1.0. Linear warmup is followed by cosine decay to one
tenth of the peak learning rate at the end of each run.

\paragraph{Information bottleneck.}
The OpenVLA implementation uses the Donsker--Varadhan MINE estimator, with an exponential-moving-average coefficient of 0.01 and three ReLU hidden layers of width 400. The critic receives pooled visual-prefix
hidden states and flattened action-token hidden states.
The policy IB coefficient is $10^{-3}$.
The critic is optimized separately with Adam at a constant learning
rate of $5\times10^{-6}$, with its loss multiplied by 0.1.

The $\pi_{0.5}$ implementation uses a Jensen--Shannon critic with two SiLU hidden layers of width 512. 
Its inputs are the masked mean of context tokens and the mean of action-expert tokens.
Marginal pairs are formed within each GPU's minibatch. The critic
maximizes the bound, while gradient reversal encourages the policy
to minimize it. The selected configuration uses an IB coefficient of
0.01 and a constant critic learning rate of $10^{-4}$.
The critic uses AdamW with the same coefficients and clipping threshold
as the policy, with one critic update per policy update.

\begin{figure*}[h]
    % \vspace{-2.3ex}
    \begin{center}
        \includegraphics[width=0.5\linewidth]{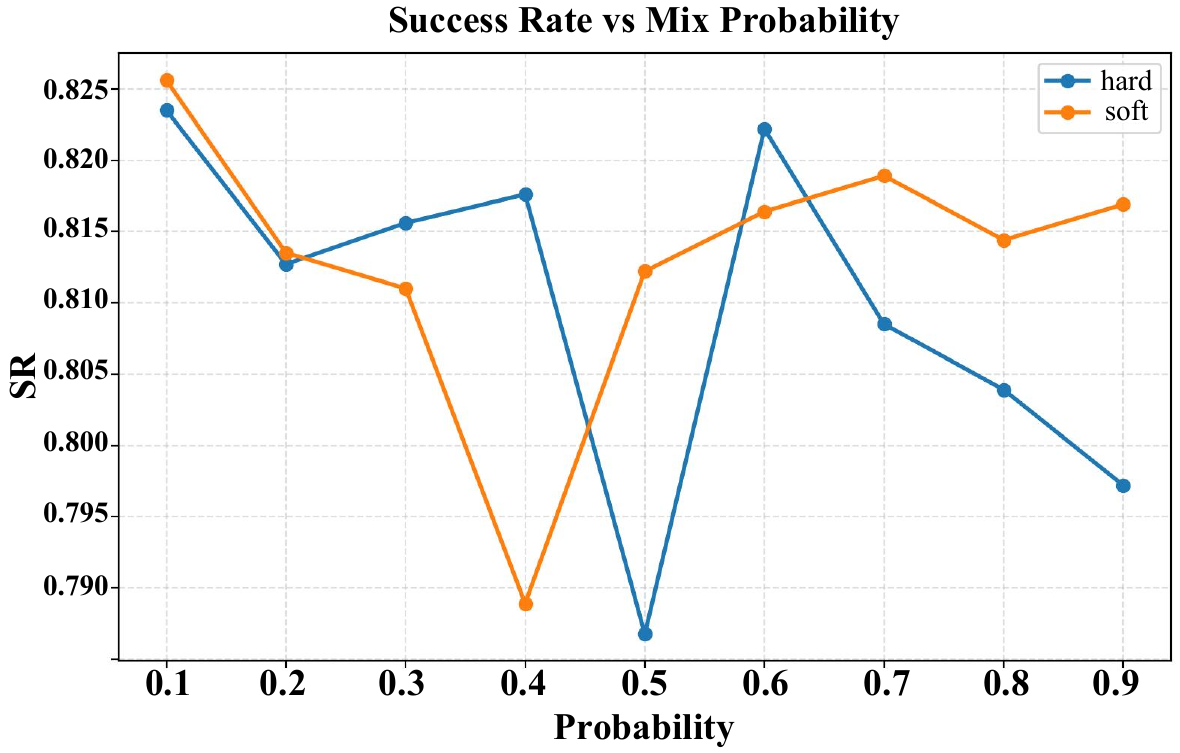}
    \end{center}
    % \vspace{-4.5ex}
    \caption{\textbf{Ablation of mixing probability.}}
    \label{fig:p_ablation}
    % \vspace{-3ex}
\end{figure*}

\begin{figure*}[t]
\centering
\includegraphics[width=0.99\linewidth]{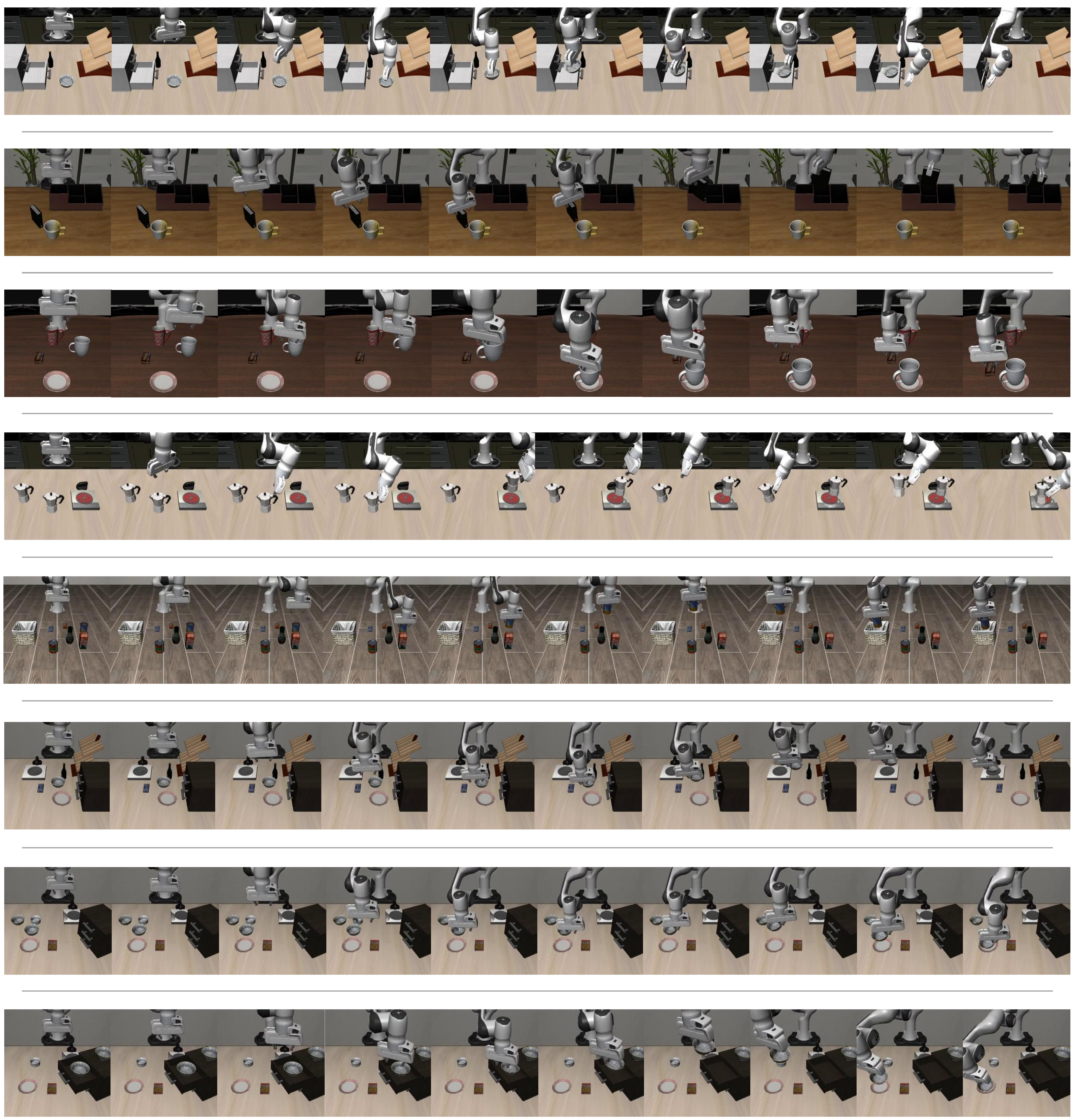}
% \vspace{-5pt}
\caption{\textbf{Visualization of manipulation trajectories on the LIBERO benchmark.}
Each row presents a sequence of key frames from a representative task execution, illustrating the manipulation progress from the initial state to task completion.
Different rows correspond to different manipulation tasks and scene configurations, demonstrating the policy behavior under the standard in-domain setting.
}
% \vspace{-10pt}
\label{fig:libero_sup}
\end{figure*}

\section{More Experiments}
\label{sec:more experiments}

\paragraph{Ablation of mixing strategies.}
% \begin{wraptable}{r}{0.52\textwidth}
\begin{table}[h]
\vspace{-5pt}
\caption{\textbf{Comparison of hard mixing and soft mixing on LIBERO-Plus.}
We compare two variants of Adaptive Mixing of Non-Invariant Information (AMI): 
\textit{hard mixing}, which performs direct interpolation between non-invariant embeddings, and 
\textit{soft mixing}, which mixes feature statistics through adaptive normalization.
Results are reported as task success rates (\%) under different perturbation types.
}
\vspace{-5pt}
\centering
\small
\resizebox{0.99\linewidth}{!}{
\begin{tabular}{ccc|cccccccc}
\toprule
 & \textbf{Hard} & \textbf{Soft} &\textbf{Camera} &\textbf{Robot} &\textbf{Language} &\textbf{Light} &\textbf{Background} &\textbf{Noise} &\textbf{Layout} &\textbf{Total} \\ 
\midrule
\multirow{4}{*}{} 
 \checkmark &             &  49.0 & 61.7 & 82.3 & 95.5 & \textbf{93.9} & 82.1 & \textbf{76.8} & 75.7 \\
 & & \checkmark & \textbf{49.6} & \textbf{63.3} & \textbf{82.6} & \textbf{96.1} & 93.2 & \textbf{82.3} & 76.1 & \textbf{76.0}\\
 \bottomrule
\end{tabular}
}
\label{tab:ablation mix}
% \vspace{-15pt}
\end{table}
We further investigate the impact of different mixing strategies used in AMI.
Specifically, we compare two variants: \textit{hard mixing}, which directly interpolates non-invariant embeddings, and \textit{soft mixing}, which performs statistic-level mixing through adaptive normalization.

As shown in Table~\ref{tab:ablation mix}, both strategies improve robustness under distribution shifts on the LIBERO-Plus benchmark.
Hard mixing introduces stronger cross-sample coupling by explicitly interpolating latent representations, which helps suppress environment-specific feature modes.
In contrast, soft mixing stabilizes feature distributions by aligning feature statistics across samples, leading to slightly better overall performance.

Overall, the results indicate that both strategies are effective for regularizing non-invariant representations, while soft mixing achieves the best average performance across perturbation types.

\paragraph{Effect of mixing strategies and probability.}
We compare hard and soft mixing on a LIBERO-Plus subset
containing 2,318 evaluation episodes, varying the mixing
probability from 0.1 to 0.9.
As shown in Figure~\ref{fig:p_ablation}, both strategies
exhibit a non-monotonic response to the mixing probability,
with pronounced performance drops at intermediate values
of 0.4 for soft mixing and 0.5 for hard mixing.
Soft mixing exhibits smaller fluctuations in the
higher-probability range of 0.6--0.9, whereas hard mixing
shows a more substantial decline.
Both strategies achieve their highest observed success
rates at a mixing probability of 0.1.
These results indicate that increasing the mixing
probability does not necessarily improve robustness.
Instead, the effectiveness of non-invariant feature
regularization depends on the interaction between the
mixing strategy and its probability, highlighting the
importance of considering these choices jointly.

\paragraph{Visualization on LIBERO benchmark.}
As shown in~\ref{fig:libero_sup}, we visualize representative manipulation trajectories of the model on the LIBERO benchmark, illustrating the policy behavior during task execution under the standard in-domain setting.

\begin{figure*}[t]
\centering
\includegraphics[width=0.99\linewidth]{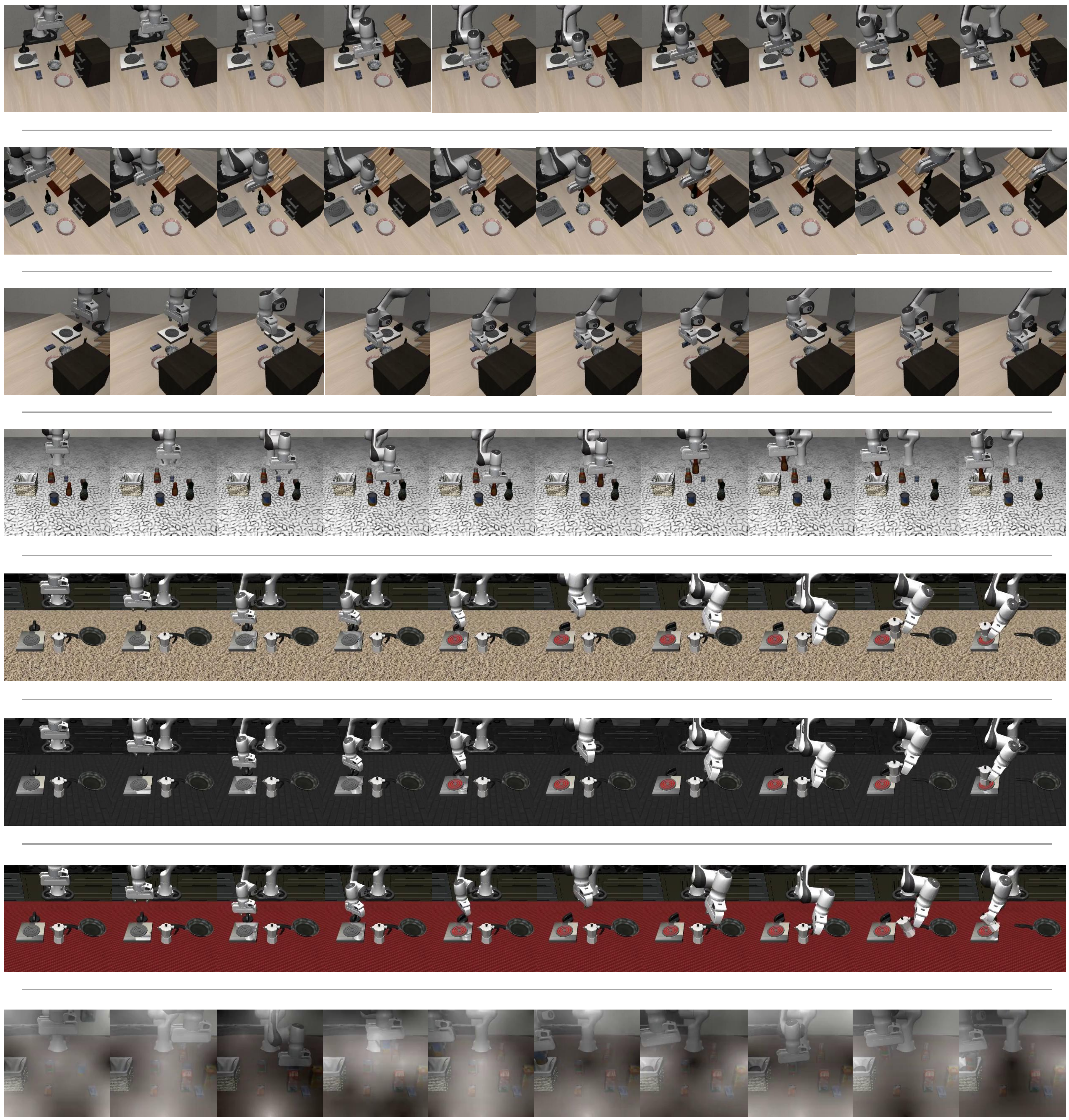}
% \vspace{-5pt}
\caption{\textbf{Visualization of manipulation trajectories under LIBERO-Plus perturbations.}
Each row presents a sequence of key frames from a representative task execution in the LIBERO-Plus benchmark.
The scenes contain various observation perturbations such as changes in background textures, lighting conditions, and visual noise.
The visualization illustrates how the policy performs manipulation under out-of-distribution environments from the initial state to task completion.
}
% \vspace{-10pt}
\label{fig:libero_plus_sup}
\end{figure*}

\paragraph{Visualization on LIBERO-plus benchmark.}
As shown in~\ref{fig:libero_plus_sup}, we visualize representative manipulation trajectories of the model on the LIBERO-Plus benchmark, illustrating the policy behavior under diverse observation perturbations.

\begin{figure*}[t]
\centering
\includegraphics[width=0.99\linewidth]{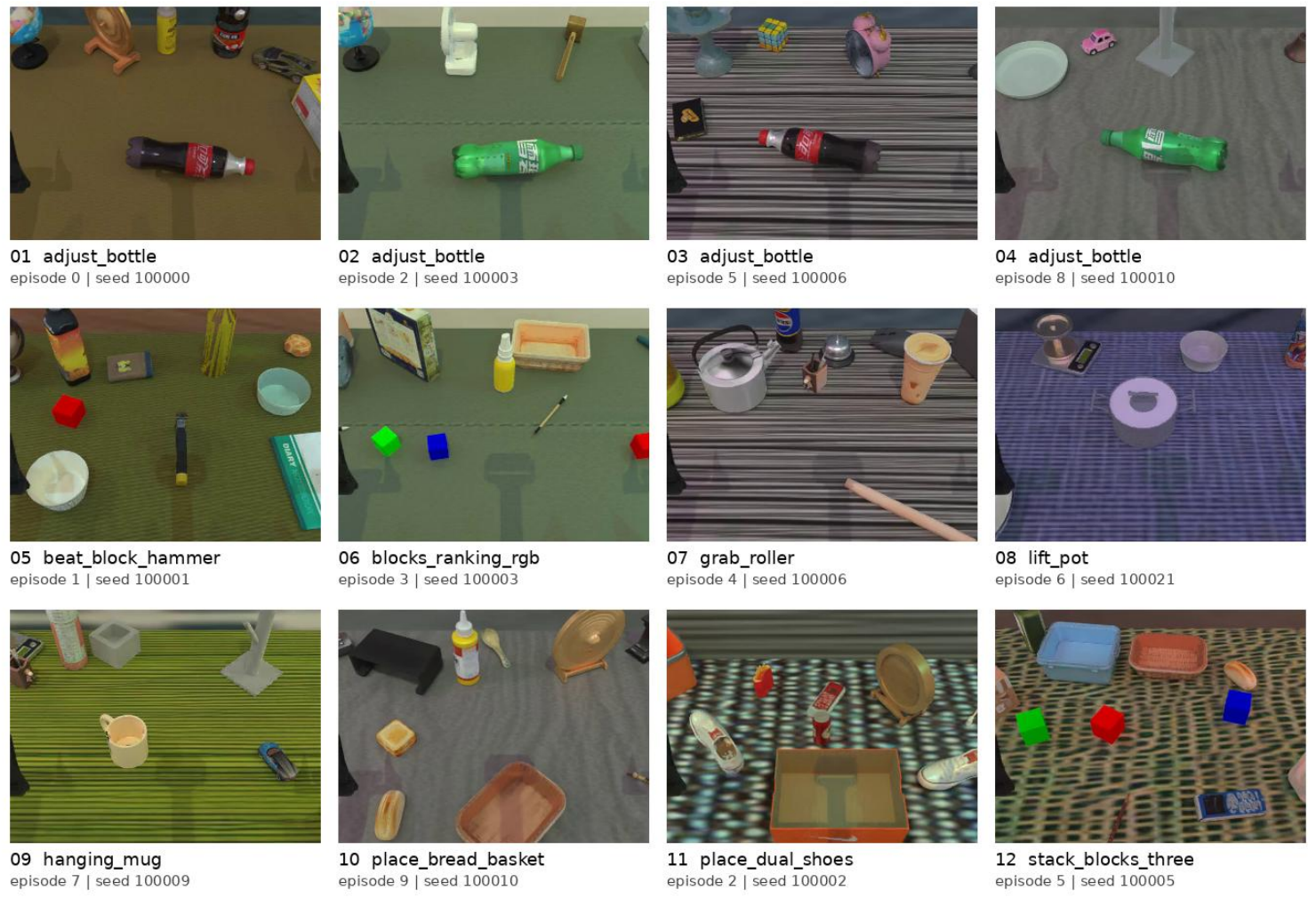}
% \vspace{-5pt}
\caption{\textbf{Visualization of RoboTwin.}
}
% \vspace{-10pt}
\label{fig:robotwin}
\end{figure*}

\paragraph{Visualization on RoboTwin 2.0 benchmark.}
As shown in Figure~\ref{fig:robotwin}, we visualize
representative observations from RoboTwin 2.0 under
the randomized evaluation setting used in our
clean-to-randomized (C2R) experiments.

\begin{figure*}[h]
\centering
\includegraphics[width=0.99\linewidth]{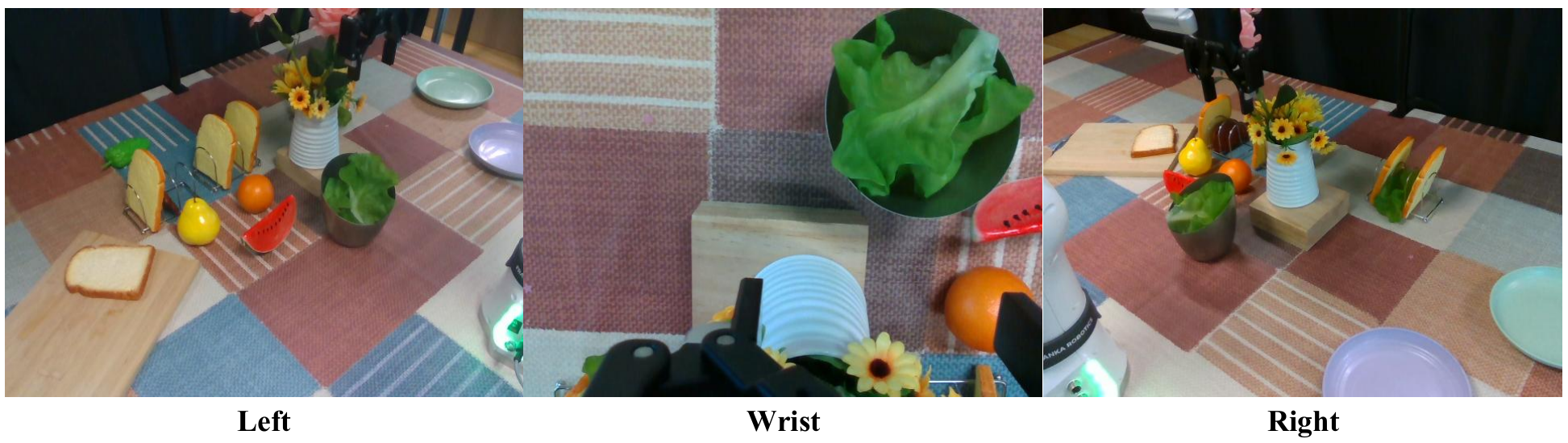}
% \vspace{-5pt}
\caption{\textbf{Three-camera observations for real-world manipulation.}
Representative views from one wrist-mounted camera and
two side-view cameras used in our Franka experiments.
}
% \vspace{-10pt}
\label{fig:3camera}
\end{figure*}

\paragraph{Real-world demo}
\textcolor{red}{\textbf{Warning.}
The middle segment of the supplementary video contains high-frequency flashing lights that may cause discomfort.
Viewers with epilepsy or sensitivity to flashing lights should avoid this segment.}

For the real-world experiments, we deploy $\pi_{0.5}$
and MixVLA on a Franka robot using three cameras:
one wrist-mounted camera and two side-view cameras.
Figure~\ref{fig:3camera} shows representative
observations from the three camera views.
The task requires the robot to pick up the sixth slice
of bread and place it on the nearest or farthest plate,
as specified by the language instruction.
We consider two lighting perturbations shown in the
supplementary video: low-frequency white flashing and
high-frequency multicolored flashing.
Each policy is evaluated over five trials per condition,
for a total of ten trials.
Under low-frequency white flashing, MixVLA succeeds
in all five trials, whereas $\pi_{0.5}$ succeeds in two.
High-frequency multicolored flashing is substantially
more challenging for both policies: MixVLA succeeds
in only one of five trials, while $\pi_{0.5}$ fails
in all five.
Aggregated across the two conditions, MixVLA achieves
a success rate of 60\%, compared with 20\% for $\pi_{0.5}$.

The final segment of the video additionally presents
our earlier implementation of the framework on
Diffusion Policy (DP), referred to as \emph{MixDP},
deployed on an xArm 6 robot.
These executions provide a qualitative illustration
of the framework's applicability to Diffusion Policy.
Our investigation was initially motivated by early
DP deployments, during which we observed frequent
execution failures when ambient lighting differed
from that during demonstration collection.
These observations motivated us to investigate training
strategies for improving policy robustness to visual
distribution shifts.

\section{Discussion}
\label{sec:discussion}

\paragraph{Limitations under Large Geometric Shifts.}

Limitations arise under perturbations that induce large geometric
changes in observations, such as significant camera viewpoint shifts.
As shown in our experiments, the improvement under the \textit{Camera}
perturbation is less pronounced compared with other appearance-related
perturbations.
This may be because viewpoint changes introduce discontinuous
variations in the observation space, where latent-space vicinal
smoothing becomes less effective.
Addressing such geometric shifts may require incorporating
explicit spatial reasoning or multi-view representations.

\paragraph{Future Research Directions.} 
Several directions may further extend the proposed framework.
First, more advanced mixing strategies could be explored to better capture the structure of non-invariant representations.
For instance, structure-aware or attention-guided mixing may provide finer control over how non-invariant information is regularized.

Second, integrating AMI with test-time adaptation or online representation updates may further improve robustness in highly dynamic environments.

Third, Self-MixVLA motivates further exploration of the
model's own optimization trajectory for improving
generalization.
Future work could investigate how to identify and reuse
parameter-update directions across training checkpoints
to construct more effective complementary representations.
Adaptive checkpoint selection and trajectory-based weight
extrapolation may help narrow the robustness gap to full
MixVLA while avoiding a separate Normal Net training run.

\end{document}